\documentclass[10pt,twocolumn,letterpaper]{article}

\usepackage{abstract}
\usepackage{cvpr}
\usepackage{times}
\usepackage{epsfig}
\usepackage{graphicx}
\usepackage{amsmath}
\usepackage{amssymb}
\usepackage{algorithm}
\usepackage{algorithmic}
\usepackage{capt-of}
\usepackage{booktabs}
\usepackage{array}
\usepackage{comment}
\usepackage{gensymb}
\usepackage{tabularx}
\usepackage{multirow}
\usepackage{booktabs}
\usepackage{siunitx}
\usepackage{multirow}
\usepackage[table,xcdraw]{xcolor}

\newcommand{\nelson}[1]{\textcolor{cyan}{}}
\newcommand{\fuyang}[1]{\textcolor{blue}{}}
\newcommand{\yasu}[1]{\textcolor{orange}{}}

\newcommand{\mysubsubsection}[1]{\vspace{0.1cm} \noindent {\bf #1}:}

\usepackage[pagebackref=true,breaklinks=true,letterpaper=true,colorlinks,bookmarks=false]{hyperref}

\cvprfinalcopy 

\def\cvprPaperID{40} 

\ifcvprfinal\fi

\title{Conv-MPN: Convolutional Message Passing Neural Network\\for Structured Outdoor Architecture Reconstruction}

\author{Fuyang Zhang\thanks{indicates equal contribution.}~, Nelson Nauata\footnotemark[1]~ and Yasutaka Furukawa\\
Simon Fraser University, BC, Canada\\
{\tt\small \{fuyangz, nnauata, furukawa\}@sfu.ca}
}

\begin{document}
\twocolumn[{
\maketitle
\vspace{-2em}
\centerline{
\includegraphics[width=\linewidth]{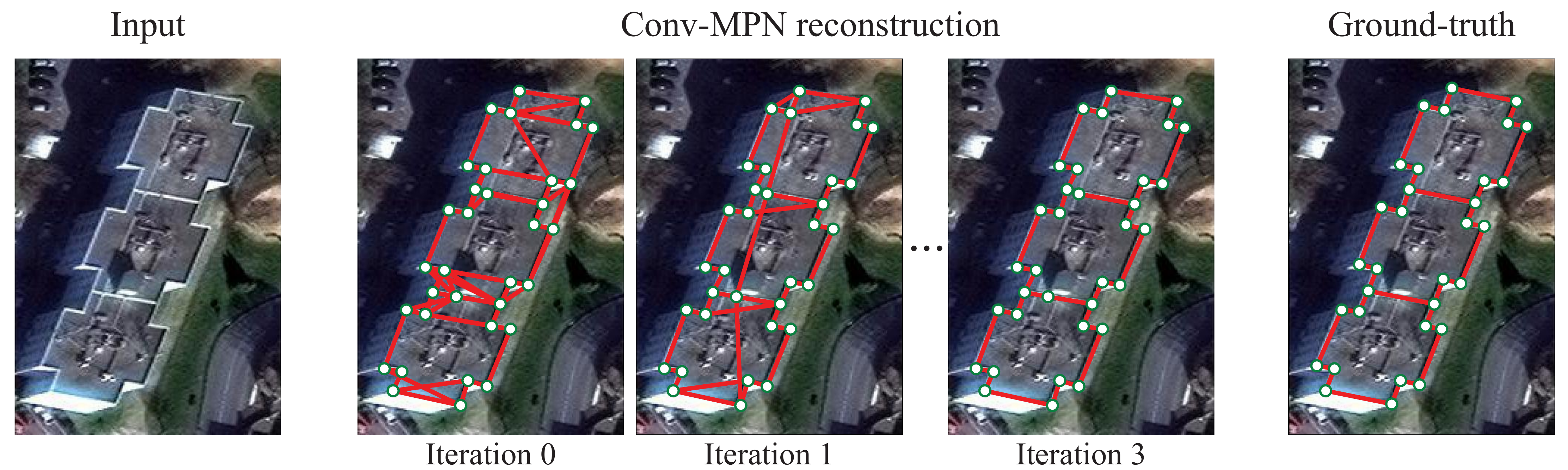}
}
\captionof{figure}{Conv-MPN, a novel message passing neural network, reconstructs outdoor buildings as planar graphs from a single image. The reconstructions after 0, 1, or 3 iterations of message passing are as shown.
%
}
\label{fig:teaser}
\vspace{1em}
}]
\saythanks

\begin{abstract}
This paper proposes a novel message passing neural (MPN) architecture Conv-MPN, which reconstructs an outdoor building as a planar graph from a single RGB image. Conv-MPN is specifically designed for cases where nodes of a graph have explicit spatial embedding. In our problem, nodes correspond to building edges in an image.
Conv-MPN is different from MPN in that 1) the feature associated with a node is represented as a feature volume instead of a 1D vector; and 2) convolutions encode messages instead of fully connected layers.
%
%
Conv-MPN learns to select a true subset of nodes (i.e., building edges) to reconstruct a building planar graph.
%
%
Our qualitative and quantitative evaluations over 2,000 buildings show that Conv-MPN makes significant improvements over the existing fully neural solutions. 
We believe that the paper has a potential to open a new line of graph neural network research for structured geometry reconstruction.
%
\end{abstract}

\section{Introduction}
Human vision evolved to master holistic image understanding, capable of detecting structural elements in an image and inferring their relationships. Look at a satellite image in Fig.~\ref{fig:teaser}. We can quickly see three building components, detect their building corners, and identify the common edges with the neighboring components.

The ultimate form of such structured geometry is the CAD representation, which enables a wide spectrum of applications such as rendering, effects mapping, simulation, or human interactions. Unfortunately, CAD model construction is still an open problem for computer vision, and is possible only by the hands of expert modelers.

Towards the automated construction of CAD geometry, the emergence of deep neural networks (DNNs) have brought revolutionary improvements to the detection of low-level primitives (e.g., corners). However, holistic understanding of high-level geometric structures (e.g., the inference of a graph) remains as a challenge for DNNs. The current state-of-the-art utilizes DNNs for low-level primitive detection,  but employs optimization methods for high-level geometric structure inference~\cite{vec20buildings, cjc2019floorsp}. Optimization is powerful, but requires complex problem formulations and intensive engineering for injecting structural constraints.

This paper seeks to push the boundary of deep neural architecture for the task of structured geometry reconstruction. In particular, we propose a convolutional message passing neural network (Conv-MPN). Conv-MPN is a variant of a graph neural network (GNN), and learns to infer relationships of nodes by exchanging messages. Conv-MPN is specifically designed for cases where a node has an explicit spatial embedding, and 
makes two key distinctions from a standard message passing neural network (MPN):
%
1) the feature of a node is represented as a 3D volume as in CNNs instead of a 1D vector; and 2) convolutions encode messages instead of fully connected layers~\cite{qi2016pointnet, qi2017pointnetplusplus} or matrix multiplications~\cite{bruna2013spectral, NIPS2016_6081}. This design allows Conv-MPN to exploit the spatial information associated with the nodes.



    %


We have demonstrated the effectiveness of Conv-MPN on an outdoor architecture vectorization problem~\cite{vec20buildings}, where the input is a satellite RGB image and the output is a planar graph depicting both the internal and external architectural feature lines.
This is a challenging problem for computer vision akin to the floorplan vectorization, which did not have an effective solution until recently~\cite{liu2017raster}. 

The main challenge lies in the inference of a graph structure with an arbitrary topology. The outdoor architecture vectorization from a satellite image is even more challenging as Manhattan assumption does not hold  due to the  foreshortening effects.
We would like to also emphasize the difference from the traditional building shape extraction problem~\cite{spacenet_challenge}, which represents a building as a set of pixels.

We qualitative and quantitative evaluate the proposed approach on more than 2,000 complex building examples in the cities of Atlanta, Las Vegas, and Paris~\cite{vec20buildings}.
Conv-MPN makes significant improvements over all the existing neural solutions.
%
%
We believe that this research has a potential to open a new line of graph neural network research for structured geometry reconstruction. Code and pretrained models can be found at \url{https://github.com/zhangfuyang/Conv-MPN}.

\section{Related work}

We first review structured reconstruction techniques based on the levels of graph structure to be inferred, then the use of message passing techniques on structured data.


\mysubsubsection{Reconstruction with a fixed topology} 
With the fixed known topology, graph reconstruction amounts to simply detecting keypoints and classifying their semantic types, because their connections are already given. Convolutional neural networks have shown to be effective in solving 
human pose estimation~\cite{newell2016stacked, xiao2018simple, sun2019deep} and hand tracking~\cite{yuan2018depth, zimmermann2017learning, xiang2019monocular}.

\mysubsubsection{Low- to mid-level structured reconstruction} DNNs detect corners and classify the presence of their connections for wire-frame parsing~\cite{wireframe_cvpr18, zhou2019end, zhang2019ppgnet}. However, the connection classification is performed for each edge independently, lacking higher level geometric reasoning that considers a graph as a whole.
%
In remote sensing, most building extraction methods represent a building as a set of pixels~\cite{hamaguchi2018building} or a 1D polygonal loop~\cite{acuna2018efficient, castrejon2017annotating, marcos2018learning, cheng2019darnet}, limiting the output to a building external boundary as a 1D loop. In contrast, we seek to infer a graph of an arbitrary topology encoding both internal and external architectural feature lines.

\mysubsubsection{Structured reconstruction (optimization)} 
The state-of-the-art graph structure inference combines CNNs and optimization, in particular, integer programming (IP)~\cite{liu2017raster,liu2018floornet,vec20buildings}. CNNs detect low-level geometric primitives (e.g., corners) or infer pixel-wise graph information (e.g., edge likelihood). IP fuses all the information and infer graph structure, which is powerful but requires complex problem formulations and intensive engineering for injecting structural constraints.


\mysubsubsection{Structured reconstruction (learning)}
A few methods learn to infer high-level geometric structure.
Ritchie~\etal~\cite{ritchie2016neurally} used DNNs to learn the arrangement of 2D strokes on a canvas with a simple shape-grammar similar to the L-system. Frans~\etal~\cite{frans2018unsupervised}. proposed an unsupervised approach to the problem.
However, their grammar is too rudimentary to represent building architecture.
Zeng~\etal~\cite{zeng2018neural} utilizes an architectural shape-grammar to reconstruct outdoor buildings from an ortho-rectified depthmap. However, their shape-grammar is again too restrictive: 1) Requiring ortho-rectification to utilize the Manhattan assumption; and 2) Modeling only small residential houses.
This paper does not rely on a shape grammar, instead learns structural regularities from examples and utilize in the structure inference.

\mysubsubsection{Message passing and convolution on graphs} Message passing has been an effective tool for high-level data reasoning~\cite{niepert2016learning, atwood2016diffusion, gilmer2017neural, kipf2016semi, defferrard2016convolutional, xu2018powerful, battaglia2018relational}.
A standard way is to extend the convolution operation over a grid of pixels to a graph of vertices~\cite{bruna2013spectral, kipf2016semi, defferrard2016convolutional, duvenaud2015convolutional, niepert2016learning}. Bruna \etal \cite{bruna2013spectral}, Defferand \etal\cite{defferrard2016convolutional} and Kipf \etal\cite{kipf2016semi} utilizes spectral analysis to define graph convolutions that act on an entire graph.
The key difference is that our convolutions do not occur in the graph domain. Conv-MPN takes a graph with an explicit spatial embedding, represent a node as a feature volume and perform convolutions in the spatial domain.
%
This framework allows Conv-MPN to exploit the spatial information associated with the nodes of a graph.

\section{Preliminaries} \label{preliminaries}
\begin{figure}
    \centering
    \includegraphics[width=\linewidth]{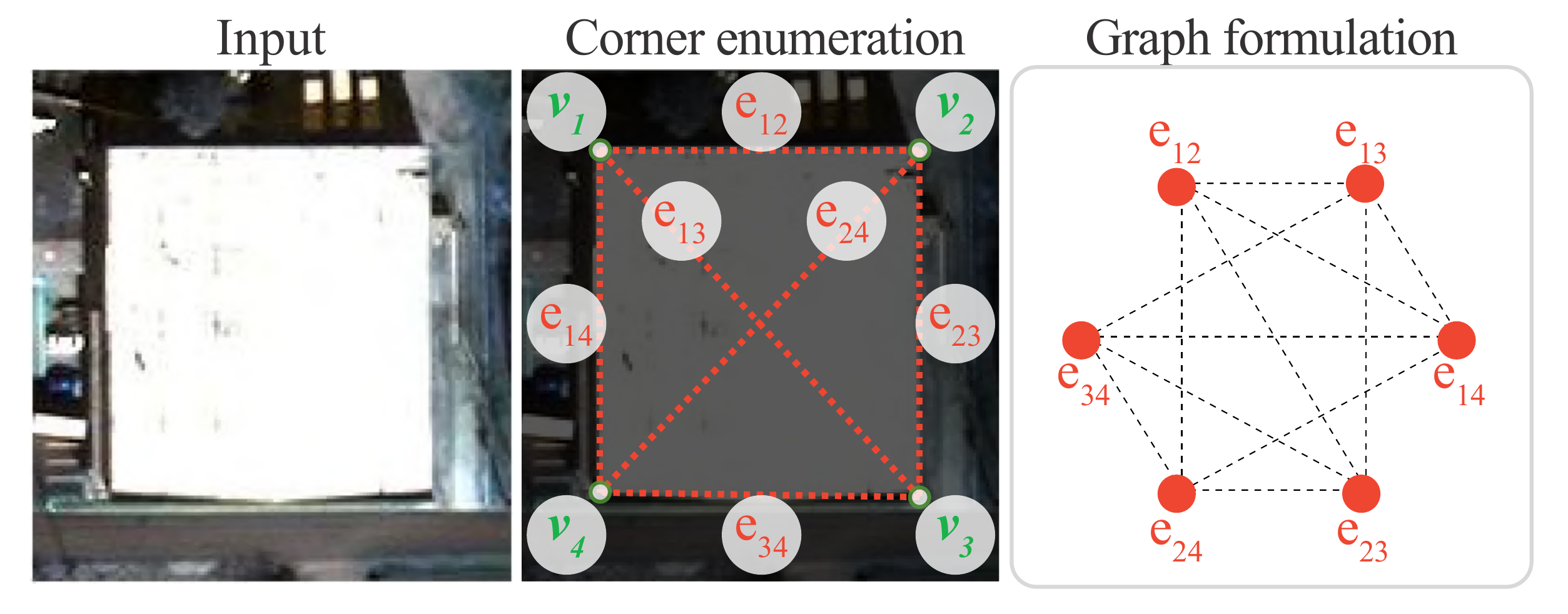}
    \caption{Preliminaries. Given a RGB image, we detect building corner candidates, enumerate bulding edge candidates, then formulate a graph for inference whose nodes are building edges.} 
    \label{fig:preprocessing}
\end{figure}

This paper tackles the 2D architecture vectorization problem from a single satellite image, where a building is represented as a 2D planar graph.
This section describes our data source and pre-processing steps (See Fig.~\ref{fig:preprocessing}).
    
    \mysubsubsection{Dataset}
    Our data source is a set of high-resolution satellite RGB images from SpaceNet~\cite{spacenet} corpus, hosted through the Amazon Web Services (AWS) as a part of the SpaceNet Challenge~\cite{spacenet_challenge}. In particular, we use an existing benchmark~\cite{vec20buildings}, which cropped 2,001 buildings into $256\times 256$ square image patches for the cities of Atlanta, Paris and Las Vegas~\cite{vec20buildings}.
    We use the same training and testing split (1601/400) as well as the metrics,
    which consists of the precision, recall, and f1-score for each of the corner, edge, and region primitives.
    Note that the satellite images are off-nadir, and buildings do not follow the Manhattan assumption due to the foreshortening effects.
    
    \mysubsubsection{Corner candidates enumeration}
    Given an input RGB image-patch $\mathcal{I}$ ($256\times256$), we use Faster-RCNN with ResNet-50 as the backbone \cite{ren2015faster} to detect corner candidates, while treating each corner as a $8\times8$ bounding box with a corner at the center. 
    %
    The model is trained using SGD with the leaning rate set to $0.0001$ and the batch-size of 1.

    \mysubsubsection{Graph formulation} \label{graph_formulation}
    Given building corner candidates, we enumerate building edge candidates by every pair of corners. Each edge candidate becomes a node in the graph for Conv-MPN inference in the next step (See Fig.~\ref{fig:preprocessing} right).
    Nodes are connected when the corresponding building edges share the same building corner.
    Notice that we have a building planar graph and the graph for Conv-MPN inference. We explicitly write ``building corner'' or ``building edge'' when terms are confusing.
\section{Conv-MPN architecture}
\label{conv-mpn}

\begin{figure*}
    \centering
    \includegraphics[width=\textwidth]{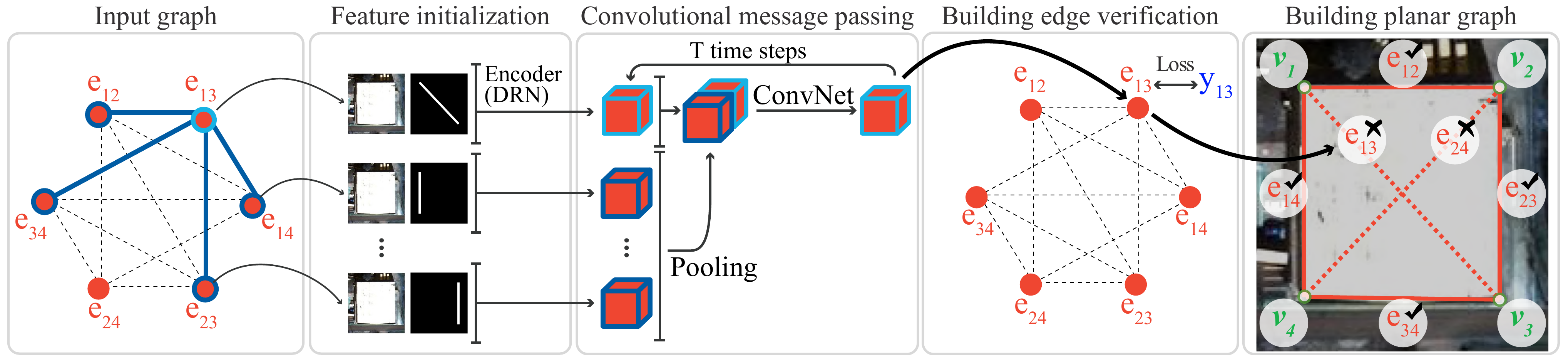}
    \caption{Conv-MPN Architecture. Given a graph, DRN encoder initializes a feature volume for each node. Convolutional message passing update feature volumes for T times.
    A building edge verification module uses a simple CNN decoder to estimate the confidence of a node (i.e., a building edge candidate).
    } 
    \label{fig:overall}
\end{figure*}

The fundamental idea behind Conv-MPN is simple yet powerful. Standard graph neural networks (GNNs)~\cite{battaglia2018relational} encode geometry information as a 1D vector numerically instead of a 3D feature volume spatially. MLP with 1D feature vectors cannot make effective geometric analysis, while convolution with 3D feature volumes enables natural spatial reasoning.
Our idea is to take the standard MPN architecture then replace 1) a latent vector with a latent 3D volume for the feature representation; and 2) fully connected layers (or matrix multiplications) with convolutions for the message encoding.
%
The section explains the Conv-MPN architecture specific to our problem setting, but it is straightforward to extend the framework to the entire GNN family.
\subsection{Feature initialization} \label{feature_initialization}
A node in the inference graph corresponds to a building-edge, which is to be represented as a 3D feature volume. We initialize the feature volume by passing a building RGB image concatenated with a binary building edge mask ($256\times256\times4$) through Dilated Residual Network (DRN)~\cite{Yu2017DilatedRN} (See Fig.~\ref{fig:overall}). More specifically, we use the first three blocks of the DRN-C-26 architecture, followed by one $3\times3$ stride 2 convolution for downsampling to $64\times64\times32$.
During training, we initialize the network parameters by the pretrained weights on the ImageNet~\cite{russakovsky2015imagenet}.~\footnote{We do not keep latent features at the graph edges unlike standard MPN for memory consideration.}

\subsection{Convolutional message passing} \label{cmp}

A standard form of feature vector update in MPN is to utilize multi layer perceptron (MLP) for encoding messages and mixing with the current feature:
\begin{eqnarray}
f_v \leftarrow \mbox{MLP}\left(f_v;  \sum_{w\in \mathbf{N}(v)} \mbox{MLP}(f_v; f_w)\right)
\end{eqnarray}
$f_v$ denotes the feature vector associated with a node $v$, $\mathbf{N}(v)$ denotes the set of neighboring nodes, and ``;'' denotes the feature concatenation.

While Conv-MPN could simply replace MLP by CNN to form a feature update rule, that would require two CNN modules and hence more GPU memory. A node feature spreads across a volume and a simple pooling could keep all the information in a message without collisions.
%
More precisely, instead of encoding a message for every pair of nodes, we just pool features across all the neighboring nodes to encode a message, followed by CNN to update a feature vector:
\begin{eqnarray}
f_v \leftarrow \mbox{CNN}\left[f_v;\underset{w\in \mathbf{N}(v)}{\mbox{Pool}} f_w \right].
\end{eqnarray}
We experimented max, sum, and mean poolings and the max pooling worked the best. We perform feature update up to 3 iterations due to the GPU memory limitation. The CNN module consists of 7 Conv-ReLU-BN blocks, which are not shared across different iterations.
%
We use Conv-MPN(t=x) to denote our architecture with $x$ iterations of convolutoinal message passing.




\subsection{Building edge verification} \label{line_verification}
After a few iterations of feature update, we put a CNN decoder to each node and output a confidence score, indicating if the corresponding building edge is true or not (See Fig.~\ref{fig:overall}). The decoder first pass the feature into a 5 Conv-ReLU-BN blocks to convert feature into $64\times64\times128$. After that, feature are downsampled into $2\times2\times128$ via max-pooling. In the end, the feature is flattened into 512 dimensional feature vector, followed by a single fully connected layer to regress a confidence score.


\subsection{Edge classification loss}
 We use the weighted binary cross-entropy loss:
\begin{eqnarray}
 \mathcal{L} = -\sum \mathcal{H}\log\hat{\mathcal{H}} - \lambda(1 - \mathcal{H})\log(1-\hat{\mathcal{H}}).
 \end{eqnarray}
 $\mathcal{H}$ and $\hat{\mathcal{H}}$ are the ground-truth and the prediction of the building edge confidence. $\lambda=3$ is used to increase the weight on positive samples.

\section{Experiments} \label{experiment}

\begin{figure*}
    \centering
    \includegraphics[width=\linewidth]{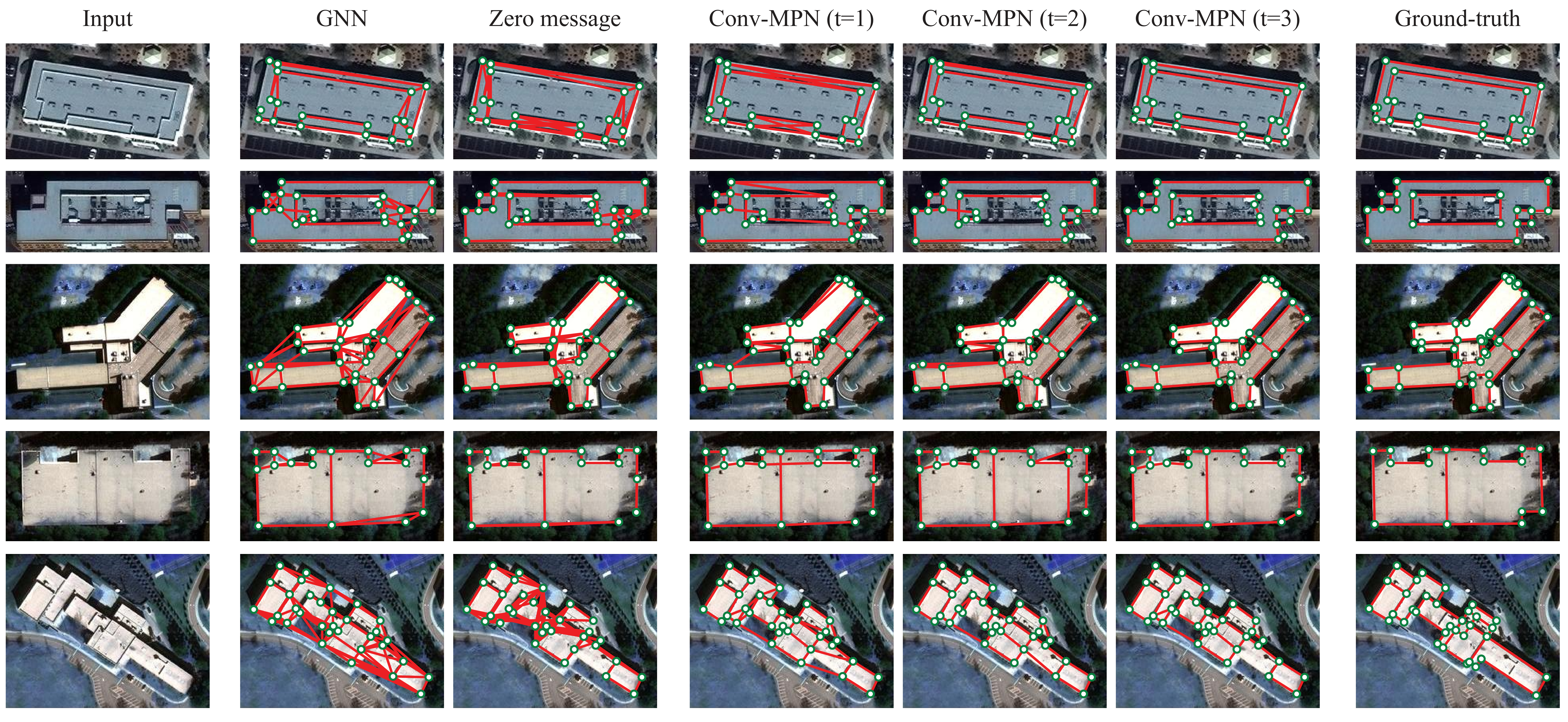}
    \caption{From left to right, an input RGB image, GNN, Zero message, Conv-MPN reconstructions after 1, 2, or 3 iterations of convolutional message passing and ground-truth.}
    \label{fig:conv_mpn_iterations}
\end{figure*}

\begin{figure*}
    \centering
    \includegraphics[width=\linewidth]{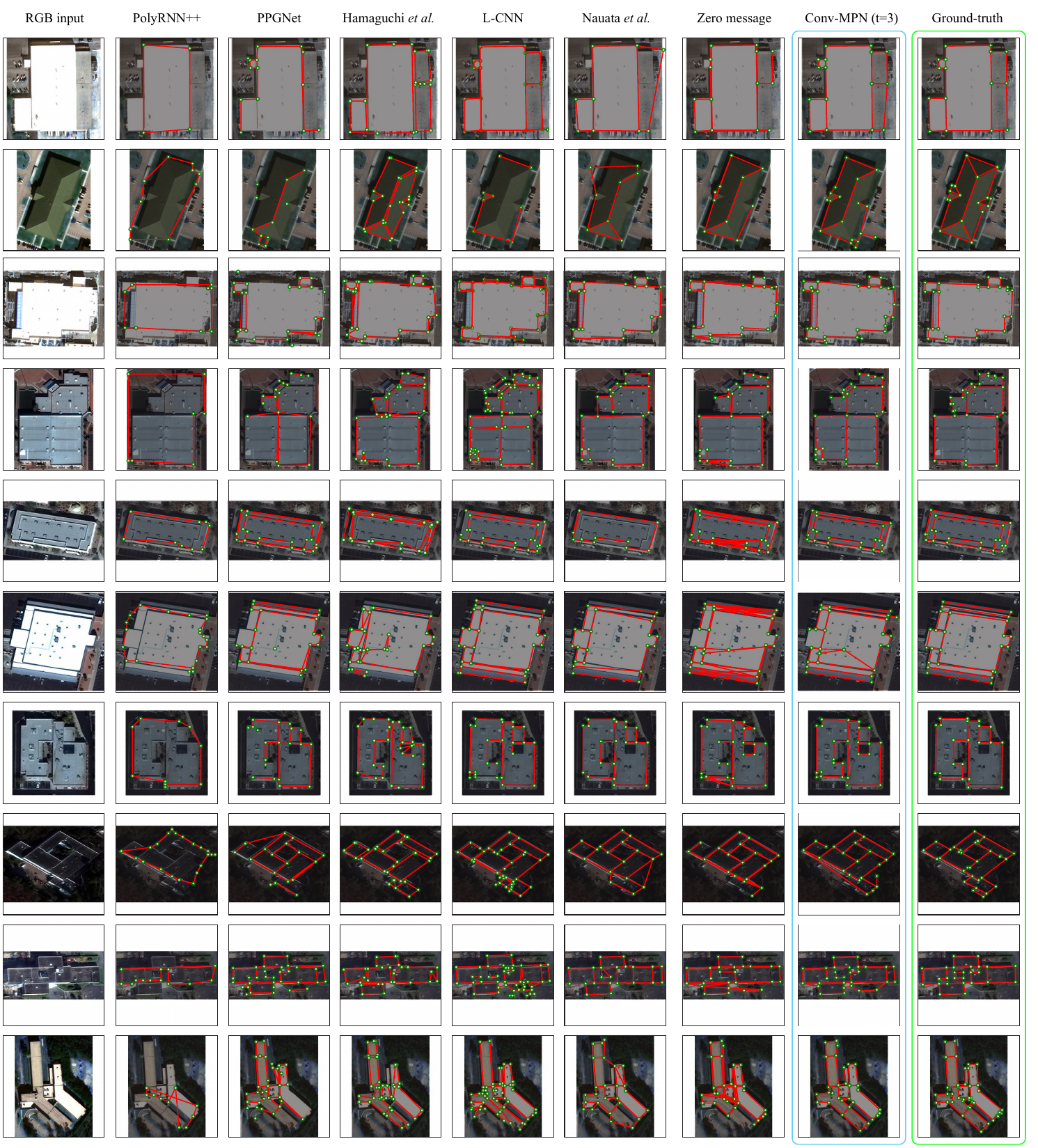}
    \caption{Comparative evaluations against competing methods. 
    PolyRNN++~\cite{acuna2018efficient}, PPGNet~\cite{zhang2019ppgnet}, Hamaguchi~\etal~\cite{hamaguchi2018building}, and L-CNN~\cite{zhou2019end} are prior-free existing methods, all utilizing DNNs. Nauata~\etal~\cite{vec20buildings} is not prior-free.
    Zero message is a variant of our Conv-MPN without any message passing. Conv-MPN is our prior-free system.}
    \label{fig:comparison_all}
\end{figure*}

\begingroup
\renewcommand{\arraystretch}{1.1}
\begin{table*}[!t]
\caption{\textbf{Comparative evaluations}: Table shows precision and recall values, when the edge confidence threshold is set to $0.5$. 
The \textcolor{cyan}{cyan}, \textcolor{orange}{orange}, and \textcolor{magenta}{magenta} colors indicates the first, second, and third best results, respectively among the prior-free methods. Nauata~\etal is the concurrent state-of-the-art method.
This method is not prior-free and uses integer optimization with hand-crafted objectives and structural constraints.}
\label{table_all}
\centering
\begin{tabular}{lrrrrrrrrr}
\toprule
Model & \multicolumn{3}{c}{Corner} & \multicolumn{3}{c}{Edge} & \multicolumn{3}{c}{Region}\\
\cmidrule(lr){2-4}\cmidrule(lr){5-7}\cmidrule(lr){8-10}
& Preci. & Recall & F1-score & Preci. & Recall & F1-score & Preci. & Recall & F1-score\\
\midrule
PolyRNN++ \cite{acuna2018efficient} & 49.6 & 43.7 & 46.4 & 19.5 & 15.2 & 17.1 & 39.8 & 13.7 & 20.4 \\
PPGNet \cite{zhang2019ppgnet} & \textcolor{cyan}{78.0} & \textcolor{magenta}{69.2} & \textcolor{magenta}{73.3} & \textcolor{orange}{55.1} & \textcolor{magenta}{50.6} & \textcolor{magenta}{52.8} & \textcolor{magenta}{32.4} & 30.8 & 31.6 \\
Hamaguchi~\etal \cite{hamaguchi2018building} & 58.3 & 57.8 & 58.0 & 25.4 & 22.3 & 23.8 & \textcolor{orange}{51.0} & \textcolor{magenta}{36.7} & \textcolor{orange}{42.7} \\
L-CNN \cite{zhou2019end} & \textcolor{magenta}{66.7} & \textcolor{cyan}{86.2} & \textcolor{orange}{75.2} & \textcolor{magenta}{51.0} & \textcolor{cyan}{71.2} & \textcolor{cyan}{59.4} & 25.9 & \textcolor{orange}{41.5} & \textcolor{magenta}{31.9} \\
Conv-MPN (t=3) [Ours] & \textcolor{orange}{77.9} & \textcolor{orange}{80.2} & \textcolor{cyan}{79.0} & \textcolor{cyan}{56.9} & \textcolor{orange}{60.7} & \textcolor{orange}{58.7} & \textcolor{cyan}{51.1} & \textcolor{cyan}{57.6} & \textcolor{cyan}{54.2} \\
\midrule
Nauata~\etal \cite{vec20buildings} & 91.1 & 64.6 & 75.6 & 68.1 & 48.0 & 56.3 & 70.9 & 53.1 & 60.8 \\
\bottomrule
\end{tabular}
\end{table*}
\endgroup

\begin{figure*}
    \centering
    \includegraphics[width=\linewidth]{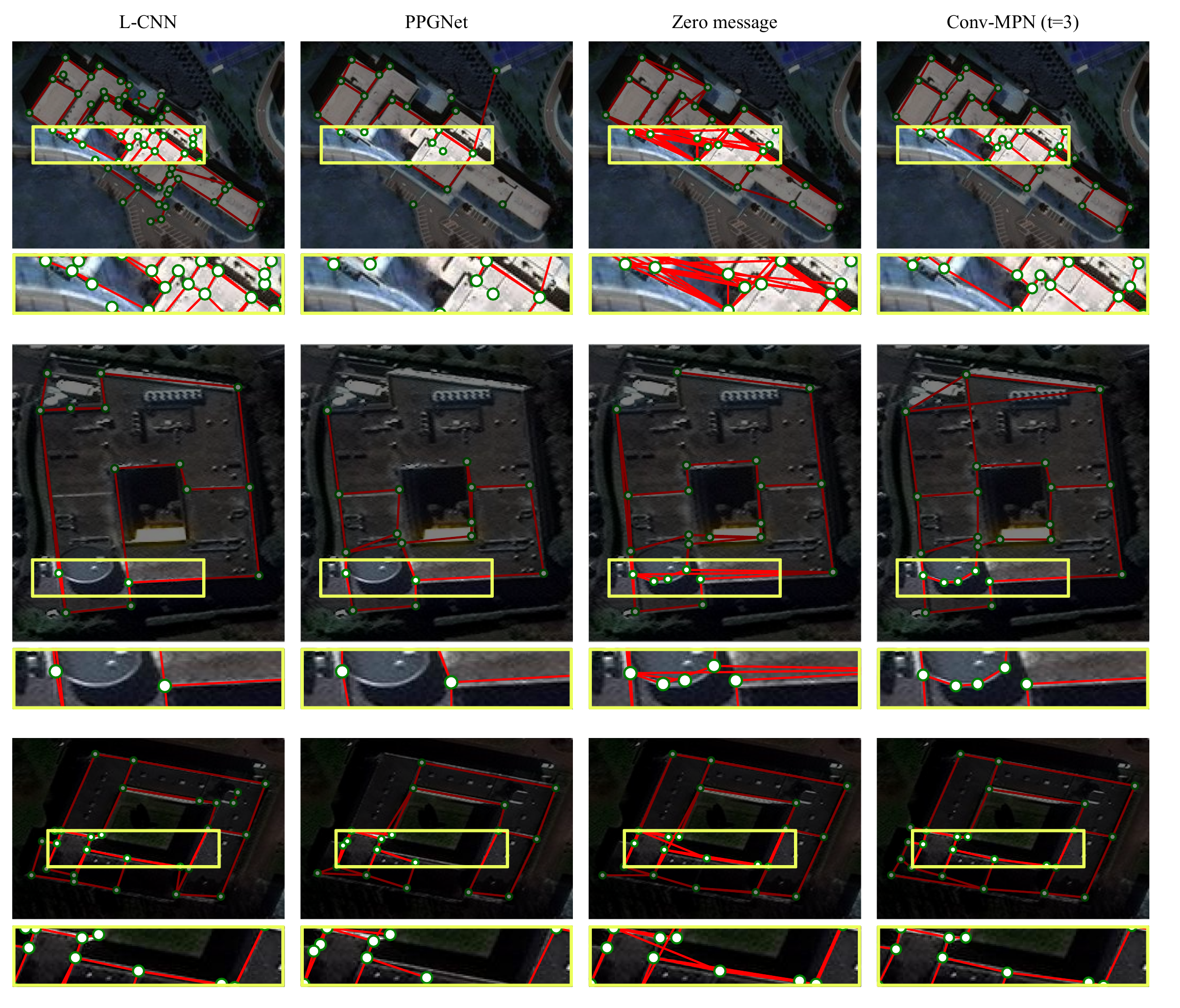}
    \caption{Close-up comparisons. From left to right, L-CNN\cite{zhou2019end}, PPGNet\cite{zhang2019ppgnet}, Zero message, and Conv-MPN(t=3). In the zooming area, we show the common mistakes that Conv-MPN can help to prevent. Typically, Conv-MPN helps removing the edge intersections, thin triangles and connecting missing edges.
    }
    \label{fig:closeups}
\end{figure*}

\begin{figure*}
    \centering
    \includegraphics[width=\linewidth]{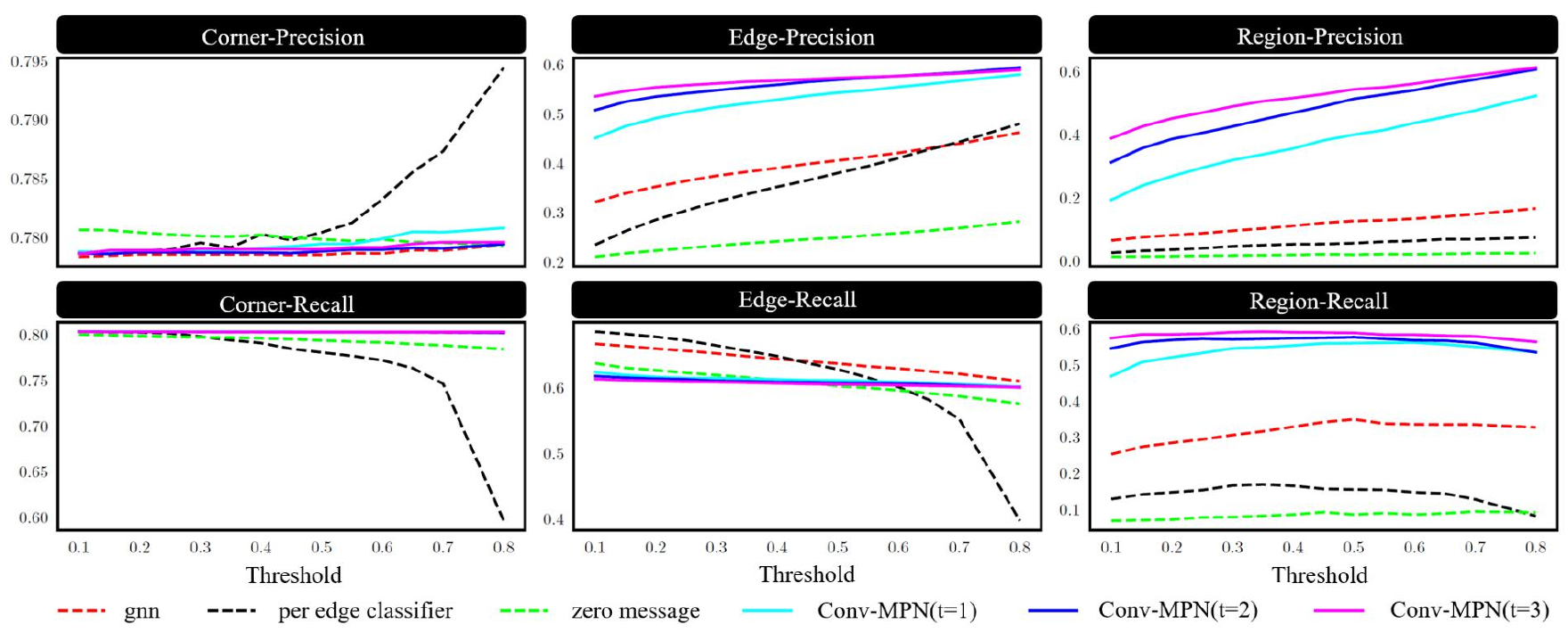}
    \caption{The precision and recall for the corners, edges and regions, while changing the edge confidence thresholds in the range $[0.1, 0.8]$ with an increment of $0.05$. We plot the precision and recall separately for clarity. Note that y-axes for different plots are not in the same scale for better visualization. 
    }
    \label{fig:pr_curve}
\end{figure*}

\begin{figure*}
    \centering
    \includegraphics[width=\linewidth]{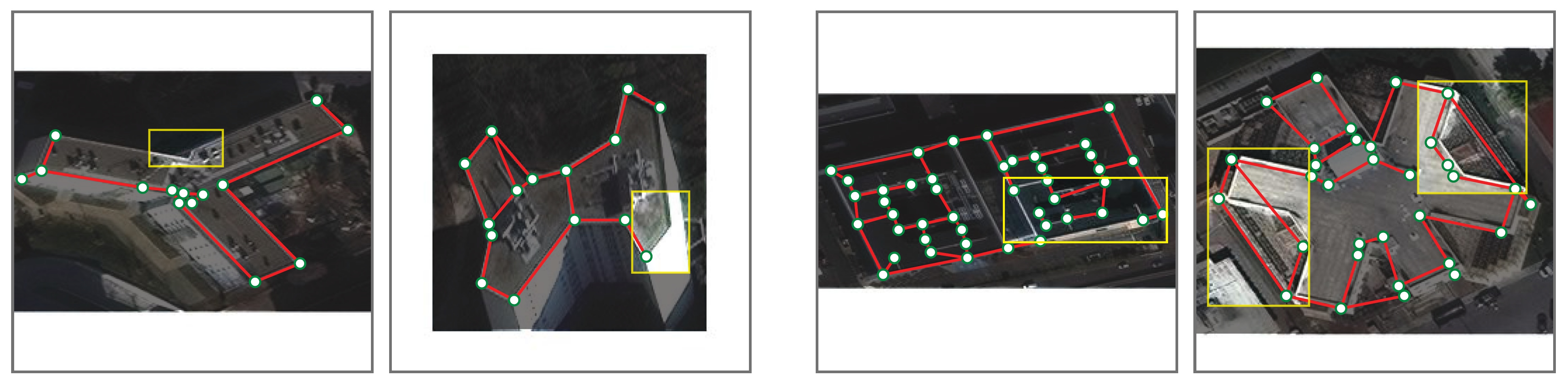}
    \caption{Failure cases. The left two examples suffer from missing corners by mask R-CNN. The right two examples show complex buildings, which Conv-MPN does not generalize well.}
    \label{fig:failure_case}
\end{figure*}

We have implemented the proposed system in PyTorch.
The learning rate is initialized to $5\times10^{-4}$, while we decay the rate by 0.8 when the testing loss does not decrease in 4 epochs. We terminate the training process when the testing loss does not decrease in 20 epochs.

Conv-MPN is GPU memory intensive due to the use of 3D feature volumes. A single NVIDIA TitanX GPU with 24G memory is used for all the experiments except for Conv-MPN (t=3), which requires two TitanX GPUs.
We set the batch size to be 1, but accumulate gradients and update the parameters every 8 batches to suppress noisy gradients.

During training, the inference graph becomes too large to fit in a GPU memory for large buildings with many building corner candidates. For training Conv-MPN (and GNN in the comparative evaluations), we used 1215 buildings that have at most 15 building corner candidates. During testing, inference requires less memory and we simply apply the trained network on large buildings, which surprisingly works well for Conv-MPN in our experiments.
Training Conv-MPN (t=1), Conv-MPN (t=2), and Conv-MPN (t=3) takes roughly 20, 30, and 40 hours. 


\subsection{Main results}
Figure~\ref{fig:conv_mpn_iterations} shows representative planar graph reconstructions by Conv-MPN. The method is able to recover complex building structure beyond the Manhattan geometry without relying on any hand-crafted constraints or priors.

Next, we conduct comparative evaluations against five competing methods: PolyRNN++~\cite{acuna2018efficient}, PPGNet~\cite{zhang2019ppgnet}, Hamaguchi~\etal~\cite{hamaguchi2018building}, L-CNN~\cite{zhou2019end}, and Nauata~\etal~\cite{vec20buildings} (See Table~\ref{table_all} and Fig.~\ref{fig:comparison_all}).
Here, we provide brief summary of the five methods.

\noindent $\bullet$ PolyRNN++ traces the building external boundary in a recurrent fashion~\cite{acuna2018efficient} and produces a 1D polygonal loop.

\noindent $\bullet$ PPGNet~\cite{zhang2019ppgnet} 
uses CNN to detect corners and classifies their connections. However, the connection (i.e., edge) classification is independent of other connections, lacking in higher level geometric reasoning.

\noindent $\bullet$ Hamaguchi~\etal~\cite{hamaguchi2018building} won the SpaceNet Building Footprint Extraction challenge~\cite{spacenet_challenge}. The method uses CNN to produce binary masks of building footprints~\cite{hamaguchi2018building}. We convert the segmentation into a polygonal loop, and use OpenCV implementation of the Ramer-Douglas-Peucker algorithm with a threshold of 10 to simplify the curve.

\noindent $\bullet$ L-CNN~\cite{zhou2019end} proposes an end-to-end neural network that detects corners and classifiers their connections. 
Like PPGNet, L-CNN also performs connection classification for each edge independently.

\noindent $\bullet$ Nauata~\etal~\cite{vec20buildings} is the current state-of-the-art for the problem, which detects 3 types of geometric primitives, classifies 2 types of pairwise primitive relationships, and uses integer programming to combine all the information into a building planar graph.


Nauata~\etal relies on integer programming with hand-crafted objectives and structural constraints.
The first 4 methods and Conv-MPN seek to learn geometric regularities or priors from examples instead.

Table \ref{table_all} shows that Conv-MPN achieves the best higher-order (region) metrics among prior-free solutions.
For the corner and edge metrics, Conv-MPN is not always the best. In particular, L-CNN outperforms Conv-MPN slightly on the edge metrics. However, as shown in Fig.~\ref{fig:closeups}, the graph structure from L-CNN is often broken, as edges are estimated independently. Figures~\ref{fig:comparison_all} and \ref{fig:closeups} demonstrate that the region metrics 
best reflect our perceptual quality of the planar graph structure, in which Conv-MPN makes significant improvements. Note that Conv-MPN stays behind Nauata~\etal~\cite{vec20buildings} on the region F1-score, which requires hand-crafted objectives and structural constraints in a complex IP optimization formulation. We would like to emphasize again that Conv-MPN learns such priors and constraints all from examples automatically, which is a phenomenal feat and makes a big improvements against all the other prior-free solutions.



\subsection{Ablation study}
We verify the contributions of Conv-MPN architecture, in particular, on the effects of 1) feature volume representation and 2) message passing. Figures~\ref{fig:closeups} and \ref{fig:pr_curve} provide the quantitative and qualitative comparisons, respectively.


\mysubsubsection{Feature volume representation}
We compare against a vanilla GNN, where we take the Conv-MPN architecture and replace ($64\times64\times32$) feature volume by a 512 dimensional vector. The feature initialization, message passing, and line verification modules are modified accordingly to match up the feature dimensions (refer to the supplementary document for the details). We conduct message passing once both on Conv-MPN and GNN for clear comparison.

Figure~\ref{fig:pr_curve} shows that GNN provides competitive results for the edge recall, but performs poorly on the other metrics. In particular, the performance gap is significant for the regions, which
requires high-level geometry reasoning and demonstrates the power of our feature representation.


\mysubsubsection{Message passing} We compare against two Conv-MPN variants that do not conduct message passing. The first variant (denoted as ``per-edge classifier'') simply does not exchange messages by cutting the inter-node connections.
The second variant (denoted as ``zero message'') is equivalent to Conv-MPN (t=1), except that it always overwrites the pooled neighbor features with a value of 0.

Figure~\ref{fig:pr_curve} shows that Conv-MPN (t=1) is superior to ``per-edge classifier'' and ``zero message'' in most metrics. In particular, the performance gap in the region metrics are again significant, indicating that Conv-MPN effectively exchanges information via the convolutional message passing.

Figure~\ref{fig:pr_curve} also shows how Conv-MPN improve reconstructions over multiple iterations of the convolutional message passing (See Figure.~\ref{fig:conv_mpn_iterations} for qualitative evaluations).
The performance improvement is consistent and strong from no iterations to 1 and 2 iterations, where per-edge-classifier can be considered as Conv-MPN (t=0). 
Due to the memory limitation, Conv-MPN (t=3) is the largest model we trained, which shows the best results, where the performance improvements start to saturate.

\subsection{Failure cases}

Conv-MPN is far from perfect, where Figure~\ref{fig:failure_case} shows failure examples. The first major failure mode comes from missing corners. If a building corner is not detected, Conv-MPN will automatically miss all the incident structure. The second major failure mode is large buildings with 30 corner candidates or more, which do not appear in the training set due to the memory limitation.

\section{Conclusion}
This paper presents a novel message passing neural architecture Conv-MPN for structured outdoor architecture reconstruction. Our idea is simple yet powerful. Conv-MPN represents the feature associated with a node as a feature volume and utilizes CNN for message passing, while retaining the standard message passing neural architecture. Qualitative and quantitative evaluations verify the effectiveness of our idea and demonstrates significant performance improvements over the existing prior-free solutions.
The main drawback is the extensive memory consumption, which is one of our future work to address. 

The current popular approach to structured reconstruction is to inject domain knowledge as hand-crafted objectives or constraints into an optimization formulation. Conv-MPN learns all such priors from examples, then infer a planar graph structure form a single image.
We believe that this paper has a potential to open a new line of graph neural network research for structured geometry reconstruction. We will share our code and data to promote further research.


\ifcvprfinal
\mysubsubsection{Acknowledgement}
This research is partially supported by NSERC Discovery Grants, NSERC Discovery Grants Accelerator Supplements, and DND/NSERC Discovery Grant
Supplement. This research is also supported by the Intelligence Advanced Research Projects Activity (IARPA) via Department of Interior / Interior Business Center (DOI/IBC) contract number D17PC00288. The U.S. Government is authorized to reproduce and distribute reprints for Governmental purposes notwithstanding any copyright annotation thereon. The views and conclusions contained herein are those of the authors and should not be interpreted as necessarily representing the official policies or endorsements, either expressed or implied, of IARPA, DOI/IBC, or the U.S. Government.
\fi

\clearpage

{\small
\bibliographystyle{ieee_fullname}
\bibliography{egbib}
}

\end{document}